# What Kind of Reasoning (if any) is an LLM actually doing? On the Stochastic Nature and Abductive Appearance of Large Language Models

Luciano Floridi[1,2], Jessica Morley[1], Claudio Novelli[1], David Watson[3]

[1]Digital Ethics Center, Yale University, 85 Trumbull Street, New Haven, CT 06511, U.S.
[2]Department of Legal Studies, University of Bologna, Via Zamboni, 27/29, 40126, Bologna, IT
[3]Department of Informatics, King's College London, 30 Aldwych, London WC2B 4BG

**Abstract**

This article examines the nature of reasoning in current, mainstream Large Language Models (LLMs) that operate within the token-completion paradigm. We explore their stochastic foundations and phenomenological resemblance to human abductive reasoning. We argue that such LLMs generate text based on learned associations rather than performing abductive inferences. When their output exhibits an apparent abductive quality – often reinforced by interface design – this effect is due to the model's training on human-generated texts that encode reasoning structures. We use some examples to illustrate how LLMs produce plausible hypotheses, simulate commonsense reasoning, and provide explanatory answers without grounding them directly in truth, semantics, verification, or understanding, and without any abductive reasoning. This duality, centred on the stochastic core of the models and the abductive appearance of the applications, has important implications for the evaluation and use of LLMs. They can help generate hypotheses and support human reasoning, but their outputs must be critically examined because they cannot discern truth or verify explanations. In the conclusion, we address five objections to our assertions, some limitations of our analysis, and provide a general assessment.

Funding: the authors did not receive support from any organisation for the submitted work.
Conflicts of interest or competing interests: conflicts of interest: he authors have no relevant financial or non-financial interests to disclose.
Ethics approval: not applicable.
Consent: not applicable.
Data or Code availability: not applicable
Authors' contribution statements: Luciano Floridi is the first and corresponding author, all remaining authors have contributed equally.

1. **Introduction: Reasoning and Generation in Context**

Current, mainstream Large Language Models (LLMs) based on the token-completion paradigm[1], like the GPT series[2] and similar systems, produce fluent language and often seem to reason, explain, and converse in a human-like manner.[3] This was true even in early versions (Floridi & Chiriatti, 2020). When users ask an LLM a question, they might receive a coherent answer with supporting evidence, as if the system had engaged in a thoughtful process of reasoning. Some have even tested this apparent reasoning ability of LLMs on complex problems in medicine or criminology and found them capable of consistently generating reasonable explanations for medical symptoms and crime scene clues. Pareschi (2023), for example, found GPT-4 effective at this type of "abductive reasoning". Results of this nature prompt us to ask: what kind of reasoning (if any) is an LLM truly undertaking? Is it genuinely following logical rules or scientific inference methods, or is it doing something fundamentally different that merely appears to be (human) reasoning? This question holds both practical importance, for assessing models' capabilities and limitations, and philosophical significance, for understanding the nature of reasoning and explanation. This is the question that the remainder of the article addresses.

Our main argument is that LLMs occupy a conceptual space "between" traditional stochastic processes and human-like abductive reasoning. On the one hand, their internal processes are entirely stochastic: during training, they gather statistical correlations from text, and during generation, they produce words based on learned probability distributions. They lack explicit representations of meaning, everyday relevance, truth values, or causality as a reasoning agent would. In fact, critics (Bender et al. 2021) have called them "stochastic parrots" to emphasise that they merely mimic language through probabilistic means, without any understanding or reasoning. Conversely, their outputs appear to share a phenomenological similarity to human reasoning. This effect is deliberately achieved through interface design, which encourages users to interpret outputs as explanations, commonsense reasoning, or analogies, but it also relates to the abductive patterns present in the data used to train the models. The result is a compelling illusion of genuine and structured inferential reasoning.

---

[1] We add this clarification to avoid any confusion. At the time of writing, alternative approaches to language modelling are emerging, such as Byte-Level Models (see the Byte Latent Transformer (BLT) developed by researchers at Meta AI), Large Concept Models (LCMs), Diffusion Models, Neurosymbolic AI systems that integrate formal reasoning with neural networks, or Selective Language Modeling (SLM). Some of them are still based on next-token prediction during the generation phase. The Rho-1 model, for example, uses SLM, which improves data efficiency and performance on specific tasks like complex math problems, though the core inference mechanism remains a form of completion.

[2] In this article, we follow the common convention of discussing the GPT series (e.g., GPT-3, GPT-4, GPT-5) to refer to the underlying core models behind ChatGPT, representing OpenAI's consumer-facing conversational AI products (e.g., ChatGPT-3.5, ChatGPT-4). Even if some of our points apply to the commercial products, we trust that the difference is clear enough to avoid confusion.

[3] We analyse text-only LLMs. Multimodal models (text-image/audio/video) may share mechanisms but introduce additional factors (e.g., cross-modal alignment), which we leave for future work.

We need to understand how stochastic processes can create outputs that resemble abductive inference, and what this implies for the broader relationship between statistical AI and human cognition and reasoning. In what follows, we explore this relationship. Section 2 defines abduction and inference to the best explanation (IBE), contrasting both with deduction and induction. Section 3 outlines statistical inference and stochastic processes and their connections to abduction. Section 4 explains LLMs' operational logic as generative models of token distributions rather than symbolic reasoning. Section 5 examines why outputs appear similar to IBE, with examples and failures, e.g., hallucinations. Section 6 considers five objections from both perspectives: either the resemblance is superficial, or LLMs exhibit weak latent reasoning. The conclusion in section 6 argues that LLMs have a stochastic core and an abductive appearance, with implications for safety and for formalising abduction. A final suggestion before we start. Sections 2 and 3 are meant to make the paper self-sufficient. Readers already familiar with abductive reasoning, inference to the best explanation, and their relationship to probabilistic reasoning may wish to skip directly to Section 4, where we turn to the specific analysis of LLMs.

## 2. Abduction and Inference to the Best Explanation

Peirce coined the term "abduction" to describe inference from effect to hypothesised cause (Peirce 1934). In a classic example, coming home to find the lawn wet, you might abduce that it rained earlier. This is not certain (someone might have run a sprinkler), but it provides a plausible explanation. Abduction thus contrasts with *deduction* (which reasons forward, in this case from cause to effect with certainty) and with *induction* (which generalises from many wet-lawn observations to a potentially probabilistic rule).

Harman (1965) later popularised the closely related idea of Inference to the Best Explanation (IBE): not only do we form explanatory hypotheses, but we also often choose the hypothesis that, if true, would best explain the evidence.[4] IBE can be understood as a form of abduction that adds a comparative evaluation step: multiple candidates are generated, then weighed by criteria such as simplicity, coherence with background knowledge, scope of explanation, and so on. The "best" explanation is then inferred as the most likely to be true. For example, if one finds footprints by the window and the laptop is missing, possible explanations might include "a burglary occurred" or "a friend borrowed it and left through the window." A reasoner employing IBE would consider which explanation makes better sense of all facts (burglary might also better explain a broken lock, etc.) and tentatively accept that one.

---

[4] Lipton's (2004) is the current cornerstone in explanation theory, elaborating the idea of IBE in depth. For more recent perspectives, see McCain & Poston (2017).

Both abduction and IBE are defeasible types of inference: their conclusions can be wrong, even if the reasoning appears sensible, because new evidence or information can defeat or invalidate them. Imagine a message from a friend apologising for borrowing the laptop. As a result, they lack the truth-preserving guarantee of deduction, but they play an essential role in everyday and scientific reasoning.

In the analytic tradition, IBE is sometimes viewed as a standalone logical rule: infer $H$ if, among competing hypotheses, $H$ would provide the best explanation for evidence $E$ if true. This can even be schematised: from $A \rightarrow B$ and observing $B$, infer $A$ as a plausible hypothesis, though not necessarily certain. The logical form is $A$ (hypothesis) implies $B$ (observation); $B$ is observed; therefore, $A$ (tentatively). Clearly, this form is not truth-preserving (many $A$s could imply $B$), but it is prudence-preserving: it suggests a good candidate to explore. If misunderstood as a deduction, it would be a fallacy. It is a fine line.

Some logicians have attempted to formalise abduction further, for example, by modelling it with conditional logic or by introducing an "explanatory" operator in modal logic. Others have presented compelling evidence from agent-based simulations that IBE trumps Bayesian inference in social settings (Douven & Wenmackers, 2017). For now, the main point is that abduction/IBE is an intuitive, narrative form of reasoning, qualitative rather than quantitative: we often explain phenomena by suggesting what circumstances could make them true, operating in the realm of possibilities and stories rather than strict deduction.

Abductive reasoning varies in strength. Sometimes, a distinction is made (Calzavarini & Cevolani 2022) between weak abduction—hypothesis generation without strong commitment—and strong abduction—inferring the most probable or best hypothesis. Weak abduction involves constructing a plausible story from the facts. Strong abduction entails choosing the best explanation among alternatives, which aligns more closely with IBE proper and may require comparative judgment or additional evidence. In cognitive science, this illustrates the difference between forming an insight and justifying it.

LLMs today seem to perform at least weak abduction: when presented with a scenario or riddle, they often generate a plausible explanation for it. They can even seem to carry out a form of strong abduction when all candidate hypotheses are explicitly provided, by selecting the most suitable one. For example, when tested on multiple-choice tasks involving abductive logical reasoning, GPT models can choose the option that best explains a given narrative. This has been shown in datasets such as the Abductive Natural Language Inference challenge, where the model must decide which of two endings best explains a story's middle (Bhagavatula et al., 2020). LLMs perform remarkably well, often at a near-human level (Balepur et al., 2024). Such findings already suggest that LLMs, despite lacking explicit reasoning, recognise patterns that align with human explanatory preferences. However, to understand why, we first need to analyse the types of processes an LLM employs, which leads us to statistics and stochastics.

### 3. Probability, Statistics, and Stochasticity

Probability theory provides the mathematical foundation for quantifying uncertainty. Instead of simply declaring a proposition true or false, probability assigns it a value between 0 and 1. The question of whether to interpret these values as subjective beliefs, inherent propensities, or limiting frequencies is a central debate in the philosophy of statistics, on which we take no position here. All we require is that probability statements satisfy the Kolmogorov axioms (1933), which provide formal rules for coherent reasoning under uncertainty. A key consequence of these axioms is Bayes' theorem, which famously describes how to update probabilities in light of new evidence.

Statistics applies probability theory to real-world data, providing methods for estimating latent parameters, performing hypothesis tests, and modelling relevant aspects of a target system. Notably, statistical inference often leads to reasoning from effects to causes, for example, inferring the impact of a drug (cause) on patient outcomes (effects) using probabilistic models and structural assumptions. In this way, statistical reasoning can justify abductive inference quantitatively. Bayesians have a formula for calculating the conditional probability of a hypothesis *h* given evidence *e*:

$$\underbrace{P(h \mid e)}_{\text{posterior}} \propto \underbrace{P(e \mid h)}_{\text{likelihood}} \times \underbrace{P(h)}_{\text{prior}}$$

The *likelihood* represents how probable the evidence would be under the hypothesis, while the *prior* encodes other relevant information, such as past results or subjective beliefs. The product of these quantities is proportional to the *posterior*, which is the main quantity of interest in Bayesian confirmation theory.

Some philosophers have argued that IBE is essentially a qualitative version of Bayesian reasoning (Lipton, 2004; Poston, 2014; Dellsen, 2024). According to this view, selecting the explanation that best accounts for the evidence simply reduces to judging which *h* maximises the posterior. Others—notably Douven (2013; 2017; 2022)—take a less conciliatory view, insisting that IBE is different from (perhaps even superior to) Bayesian updating, since it violates the principle of conditionalisation by conferring a bonus on preferred explanations that is not encoded in *P(e|h)*

or *P(h)*. Frequentists, meanwhile, dispense with priors altogether and focus instead on maximising likelihoods and controlling error rates.[5] Though philosophical disputes between these various statistical camps are rich and occasionally heated (Mayo, 2018), they tend to produce similar results in many applied settings, especially when datasets are sufficiently large.

A data-generating process that includes random variables and/or probabilistic transition rules is said to be "stochastic". A stochastic process, such as a coin toss,[6] is inherently random—though not necessarily in an unconstrained way. Outcomes may vary within a narrow band, for example, if we have a 95% probability of seeing between 45 and 55 heads in 100 tosses. Although individual outcomes remain unpredictable, we make informative and testable claims about the long-run behaviour of a stochastic process. Note that whether a system is deterministic or stochastic can depend on the level of description and choice of input variables. If Jones wears a suit on all and only those days when his morning coin toss comes up heads, then his clothing choices may appear random to the outside observer. Of course, conditional on the coin toss, Jones is an automaton, at least in sartorial matters. Similarly, seemingly random aspects of a machine learning algorithm, such as the initial values of weights or which samples are selected for a given batch update, are in fact deterministic functions of a seed parameter that can be set at the top of a coding script.

In practical reasoning and AI, probabilistic methods have shown strong capabilities for abductive tasks. For example, Bayesian networks can be used to compute the most probable explanation for observed symptoms in medical diagnosis, effectively performing IBE by maximising posterior probability. Machine learning classifiers can be trained to perform similar tasks, as exemplified by numerous high-profile examples. Since LLMs are optimised to predict text, one might also suspect that they engage in a form of probabilistic inference, although over token sequences rather than explicit scientific hypotheses. We will examine this connection shortly.

First, however, it is crucial to recognise that probabilistic inference often aligns with abductive reasoning in scientific discovery and everyday thinking. Reichenbach (1938) and subsequent philosophers of science described inference as comprising two parts: the *context of discovery*, where abduction or IBE generates hypotheses; and the *context of justification*, where we test those hypotheses, almost always via statistical inference. This two-stage model is simple but effective: abduction provides the candidate, and induction assesses it. Interestingly, LLMs seem to perform only the first part. They generate candidates (explanations, answers) but do not genuinely validate them against reality (unless they are specifically augmented by other systems, which only proves the point). They aim to model the conditional distribution of tokens in text, not to evaluate truth.

---

[5] In fact, frequentists arguably encode prior beliefs in their decision rules for rejecting null hypotheses. For example, an extremely low Type I error rate $\alpha$ indicates a demand for extraordinary evidence to reject the null—presumably because we assign it a high prior. See Lehmann & Romano, (2005, Ch. 3).

[6] With sufficient detail about initial conditions, air dynamics, etc., we could in principle render coin tosses deterministic. The reader is welcome to substitute some other example here if they prefer, e.g. radioactive decay.

In statistical terms, LLMs perform prior predictive sampling but lack an external feedback loop for posterior evaluation. We will revisit this limitation when discussing their tendency to hallucinate plausible but false information. For now, the main point is that statistics and probability provide a quantitative framework for abduction, offering formal systems to assess the likelihood or explanatory power of a hypothesis.

The stage is now set to examine LLMs. How does an LLM generate its outputs, and why do those outputs often have an explanatory or inferential nature? We now turn to this question.

## 4. LLMs as Stochastic Engines of Text

LLMs mainly work through statistical inference over language data. During training, an LLM processes enormous amounts of text and optimises a model (usually a neural network transformer)[7] to predict the next token (word or sub-word) based on the preceding context (Vaswani et al. 2017). The result is essentially a complex probability distribution: for any particular sequence of tokens/words, the model can assign likelihoods to potential continuations. When generating output, the LLM takes a user prompt as input and samples from this learned distribution to produce the next word, then repeats the process to continue. The simplest method of generation is to select the highest-probability next token each time (greedy decoding), which produces the most likely continuation. Usually, some degree of randomness is introduced through temperature and sampling, allowing varied and creative responses rather than a single, fixed answer.

Regardless of the approach, the process remains stochastic: either inherently (with sampling) or effectively (since training involves discovering a model that encodes frequencies and correlations from initial random weights). As Shanahan (2022) explains, when we prompt an LLM with a question like "Who was the first person to walk on the Moon?", we are not directly accessing a knowledge base or reasoning about the Moon landing. In reality, we are asking: given the statistical distribution of words in its training data, what is the most likely continuation of the prompt "The first person to walk on the Moon was…"? The model outputs "Neil Armstrong" because that is the most statistically common completion in its training data for that sentence prefix. Essentially, an LLM functions as an "engine of stochastic inference": it takes input and produces an output that is most statistically likely given that input and its internal model, which encodes a vast number of inferred statistical relationships.

---

[7] A "transformer" is a stack of self-attention and feed-forward layers. Tokens (e.g., words) are mapped to embeddings, positional information is added, and causal self-attention lets the model weight earlier tokens when predicting the next one. In effect, the transformer parameterizes the conditional probability: its learned weights store the correlations the model later samples from during generation.

This process lacks explicit logical rules, deliberate hypothesis testing, or reference to an external world model. It is driven solely by data correlations. The compelling aspect is that this stochastic process can produce outputs that closely resemble deliberate reasoning. Why does this happen? Setting aside interface tricks, we can focus on two main factors: (1) latent knowledge and (2) emergent pattern completion.

First, through exposure to billions of words, an LLM acquires a broad range of information about the world. It "knows", in a statistical sense, many facts, relationships, and even commonsense truths, simply because these are reflected in language use. It also learns common patterns of explanation and argument, such as how "because" often introduces an explanation, and that scientific questions are answered with specific explanatory forms. This latent knowledge allows the LLM to retrieve relevant information in response to questions. For example, ask it "How do jellyfish reproduce?" and it will likely generate a description of jellyfish life cycles. It does not search a biology database; instead, it has absorbed many textual descriptions about jellyfish, and the phrase "jellyfish reproduce by..." statistically leads to those descriptions. If we trained an LLM solely on astrological data, it would produce astrologically plausible answers.

Second, pattern completion can simulate reasoning steps. If a chain of reasoning often solves a problem in a text, the LLM may generate such a sequence. A notable improvement is that prompting LLMs with "let's think step by step" often leads them to produce a logical chain of thought, which enhances accuracy on multi-step problems (Wei et al. 2022; Kojima et al. 2022). The model is not suddenly performing real deduction; instead, the prompt triggers an output mode that mimics how humans outline reasoning steps, which strongly correlates with correct solutions in the training data.

The ongoing debate is whether LLMs merely reproduce surface patterns or possess some form of implicit models of the world and inference abilities. Some researchers argue that LLMs develop an implicit world model and can perform limited reasoning within it, thus exhibiting emergent reasoning as scale increases. Others maintain that any reasoning success is simply a superficial pattern-matching trick and would fail with slight variations in problems, hence calling successes "luck" or artefacts (Balepur et al., 2024). For example, one study (Webb et al., 2023) found that GPT-3 could solve some specific analogy puzzles as well as humans, leading to claims of emergent analogical reasoning. However, later analysis suggested these successes might not indicate flexible analogical reasoning, as slight modifications to the task or content can cause the model to fail, implying it has not truly captured the underlying relational reasoning. Consequently, the literature shows inconsistent findings: some report impressive logical feats by LLMs, while others highlight brittle failures or reliance on spurious cues (Bang et al., 2023).

What is clear is that LLMs lack specific abilities that human reasoners have. They do not understand the text they generate in the way humans assign meaning; they lack grounded semantics connecting words to the physical world or perceptual experiences (Harnad 1990, Harnad 2024). They also do not possess an inherent concept of truth or verification beyond what their training data provides. The "stochastic parrots" metaphor highlights two limitations: (a) LLMs are limited by their training data; they can remix, rephrase, and build on the data, and can be creative, but if the data contain factual gaps or biases, so will the model;[8] and (b) LLMs do not know whether they are right/correct or wrong/incorrect. A consequence of (b) is that they cannot lie in the ordinary sense in which, for example, Iago lies to Othello, that is, by making statements that they *believe* to be false (one must have an *understanding* of what counts as true or false, but one may still lie by accidentally telling a truth that one believes to be a falsehood), with the *proactive, conscious,* and *motivated intention* to deceive or manipulate others.[9] An illustrative example is the phenomenon of AI "hallucinations," in which an LLM invents a non-existent source or confidently offers a fabricated statement or explanation. For instance, when asked about a historical figure's cause of death, the LLM might create a plausible narrative if it cannot recall the fact, because providing any answer with an authoritative tone is statistically more likely than stating "I don't know" (especially if training data rarely include the AI saying it does not know). This tendency shows that the abductive style of LLM outputs is a double-edged sword: the model proposes an explanation or answer because that is what fluent, human-like responders do, and because models are trained to be 'helpful', projecting certainty so as not to undermine their perceived credibility (Yin, Wortman Vaughan and Wallach 2019). However, unlike a human expert, current LLMs do not have direct perceptual or embodied access to the world, nor do they possess conscious mental states. They can simulate expressions of positionality and uncertainty in language, but these are not grounded in lived experience; their reliability depends entirely on training, calibration, and system design, rather than on human-like understanding (Kerr et al. 2022). Any connection to real-world evidence must be deliberately engineered, as in retrieval-augmented systems, which provide external access to information rather than embodied grounding. The result can be a convincing but entirely incorrect answer, essentially a confabulation (Ji et al., 2023). Human abductive reasoning can also mislead us—scientists throughout history have hypothesised elegant explanations that proved false—but humans have additional safeguards, like new evidence, experiments, logical scrutiny, debates, and even LLMs themselves. LLMs, unless enhanced with tools or human oversight, currently lack these safeguards by default.

---

[8] For example, Yuan (2023) proved that a foundation model can solve a downstream task by prompting alone if and only if the task is representable in the category induced by the pretext task; otherwise prompting is provably insufficient, though fine-tuning may succeed.

[9] We clarify this in order to avoid anthropomorphic confusions between "lying" (which describes a state of the sender of the message) and the possibility that outputs by LLMs may mislead or deceive (which describes the effects of a message on its receiver), see for example Hagendorff 2024. For this reason, the literature has drawn on Frankfurt's concept of 'bullshitting' to characterize specific errors made by LLMs, on the grounds that such models often generate outputs indifferent to their truth conditions rather than intentionally deceptive, see for example Hicks et al. 2024.

LLMs seem to perform a kind of zeroth-order abduction (Kojima et al. 2022): given a prompt, they generate a plausible continuation (a hypothesis or explanation) based purely on learned associations. In reality, their operation is driven by maximising the probability of the sequence, which OpenAI researchers term "next-token prediction as the core objective". The model does not understand what an explanation is, but it produces text that follows the typical phrasing and structure of explanations. It does not reason about causes from scratch but outputs typical causes for typical effects observed in the training data. That is why a well-trained LLM can surprise us with accurate answers and even creative explanations. LLMs have effectively absorbed patterns of human abductive reasoning as expressed in writing. However, when faced with inputs that go beyond their training (truly novel situations or complex multi-step logical puzzles), the facade can crack.

**5. The Phenomenology of Plausibility: Why LLMs Feel Like IBE**

When users interact with an LLM-based AI, such as a chatbot or assistant, they often perceive the AI's responses as if they were created by an intelligent mind reasoning through the question. The AI's answer "makes sense": it addresses the question with relevant points, sometimes even providing justification or analogies. This phenomenology of plausibility can be pretty compelling. It explains why people have attributed understanding and even sentience or consciousness to advanced chatbots. What underpins this phenomenology? In large part, it is because the LLM's training on human language enables it to mimic how humans communicate explanations and reasons. Human-written text in its training data often results from IBE. For example, many Wikipedia articles, Q&A forums, or scientific papers present evidence and then offer an explanation or conclusion. The model has absorbed these patterns. Therefore, when prompted to explain something, it generates a response that not only states a fact but often justifies it, following a structure like "We observe X; a plausible explanation is Y, because...". It is likely to include causal connectives ("because", "thus", "therefore") and explicit reasoning steps, because that is how explanations are typically structured in the training data. From the user's perspective, it genuinely feels as if the model has reasoned to that answer.

For example, a user asks: "Why might my car not start on a cold morning?" The LLM might respond: "It could be due to a weak battery, as cold weather reduces battery efficiency, making it harder to deliver the necessary current. Another possibility is thicker engine oil at cold temperatures, which makes the engine harder to turn over. Based on your description, the battery is the most likely explanation". This answer mimics that of a mechanic or knowledgeable friend performing IBE: listing hypotheses and selecting one as the most plausible. Yet, the LLM lacks actual understanding or mental grasp of cars; it strings together probable sentences about car troubles. It has "seen" numerous explanations of car failures in its training data and reproduces that pattern. Crucially, the final sentence, "Based on your description, the battery is most likely", is something it has "learned" as a typical conversational move (people often conclude with the most likely cause). The answer's convincingness stems from its alignment with known causal mechanisms (batteries and temperature) and from the knowledge it acquired from its training data. Because the explanation is consistent with common sense, users tend to accept it as reasonable. Essentially, the LLM manages to produce the same explanation a human reasoner would likely choose. In many everyday situations, the "best explanation" is obvious (e.g., "car not starting in cold + battery issues" is a common trope). LLMs excel in these scenarios because they echo the obvious, common explanations.

However, in less common situations, LLMs can falter or produce a confident-sounding explanation that is subtly incorrect. For instance, consider a medical diagnostic scenario in Pareschi's study: the LLM is given patient symptoms that are somewhat unusual. A recent version of GPT-x might suggest a diagnosis that seems to fit, perhaps a rare disease mentioned in its training data. It then provides reasoning: "Symptom A and B together could indicate Disease Y, because Y is known to cause both". If that disease is actually known and plausible, the explanation appears convincing. But if the correct diagnosis is something the model has not strongly associated with those symptoms—maybe a very rare condition or a novel combination—it might overlook it and stick to something more obvious but wrong. Unlike a human doctor, who carefully weighs evidence (or at least can and should), the LLM "does not know what it does not know"—it has no awareness of its own ignorance—nor does it necessarily detect subtle inconsistencies. It might even invent an explanation if none readily comes to mind—meaning, if none is strongly embedded in its weights. For example, models can create fictitious medical syndrome names that sound plausible, cite "studies" that do not exist to support their explanations, or invent body parts that do not exist.[10] These are clear signs of simulation without verification. The model understands the form of an explanation—what technical terms and justifications should appear—but does not ground it in facts or literature. This problem is compounded by the recently observed behaviour of "sycophancy",[11] another unfortunate, anthropomorphic term. This is the tendency of LLMs to generate outputs that prioritise alignment with user beliefs or preferences over factual accuracy. Because users frequently prefer convincingly-written sycophantic responses to correct responses, they are not minded to 'fact-check' if the output supports their explanation (Sharma et al. 2023). In terms of IBE, it is as if the model always chooses an explanation, even when none is justified—it cannot "resist" explaining because generating a plausible and preferable continuation is its task. This could be termed over-abduction: a human reasoner might say, "I'm not sure; more information is needed", while the LLM often makes a guess regardless.

---

[10] See https://www.theverge.com/health/718049/google-med-gemini-basilar-ganglia-paper-typo-hallucination

Nonetheless, this phenomenological similarity to human reasoning can be put to positive use. A notable application is assisting human judgment: LLMs can generate hypotheses that a person might not have considered, effectively broadening the scope of abductive search. For example, in scientific research, one could ask an LLM to suggest possible explanations for an experimental anomaly. It might propose several—drawn from analogous cases in literature or general scientific knowledge—some of which could be genuinely insightful. In this way, the LLM functions as an abduction generator, supporting the human reasoner during the discovery phase. It then becomes the human's task to carry out the justification phase, evaluating those hypotheses and developing the correct reasoning. Experiments in human–AI collaborative reasoning (Zhou et al., 2024) indicate that LLMs can provide creative inputs or initial explanatory hypotheses, albeit often mixed with irrelevant suggestions. In essence, LLMs function like brainstorming assistants that toss out ideas without filtering for quality. After all, they work like statistical interfaces to an enormous amount of data accumulated for millennia by generations. A cautious human collaborator can sift through and assess them. In one study already mentioned (Pareschi, 2023), GPT-based assistants generated multiple hypotheses about how a crime might have occurred; the investigators found that some hypotheses were novel and worth exploring, even if ultimately incorrect. Here, verification—the very trait that LLMs lack—is supplied by the human, while their broad associative knowledge complements the human's narrower focus. This synergy suggests that LLMs, despite not reasoning on their own, can enhance human reasoning processes when used carefully. To adapt a common metaphor, they are the new box that can help think outside the old box. Philosophy of mind and cognitive science might see this as outsourcing part of the cognitive labour involved in hypothesis generation to an artificial agent. This artificial agent achieves this through stochastic pattern matching over the corpus of human culture, which differs significantly from how a human brain reasons. Yet, the final product (the hypothesis) might be similar or even identical and hence indistinguishable at the output. This raises important questions. In particular, if an AI can generate the same explanatory hypothesis a human would, does it matter that the process was different? From an epistemological standpoint, perhaps yes—justification is significant—but regarding the content of the hypothesis and our interpretation of it, maybe not. The example dialogue (see Example 1 below), created for this paper, can serve as a tangible illustration of LLM abductive behaviour.

---

[11] See https://openai.com/index/sycophancy-in-gpt-4o/.

> A user presents a small riddle to an LLM: "Alice found Bob unconscious on the kitchen floor next to a spilt cup of coffee. What might have happened?"
> The well-performing LLM might respond: "It's possible Bob slipped on the spilt coffee and hit his head, causing him to lose consciousness. The coffee cup could have fallen from his hand or a counter, creating the spill. Another possibility is that Bob fainted (perhaps due to a medical condition or the heat of the coffee) and dropped the cup as he fell. Given that the coffee was spilt, the scenario of slipping seems plausible; the spill could be both cause and result of the fall."

Example 1 – A first dialogue with an LLM

This answer in the example clearly shows weak abduction (coming up with hypotheses) and even an attempt at strong abduction (judging slipping as more likely based on the evidence). The LLM does not know which is correct but copies the reasoning a human investigator might use. It seems logical to the user. Indeed, it might be right. However, it could also be spinning a narrative that happens to fit common patterns (people slipping on spills are a familiar story). If Bob actually had a diabetic fainting spell, the model's preferred hypothesis might be wrong. Here, the risk is trusting the model's explanation too much. As a second example (Example 2), consider analogical reasoning, which can be linked to abductive reasoning (both involve matching a situation to a known pattern to explain it).

> A user asks an LLM: "The brain is like a computer in what way?".
>
> The well-performing LLM might respond with an analogy explanation: "Both the brain and a computer process information and have different components handling different tasks (like brain regions vs. computer hardware). The brain's neurons fire electrical signals, analogous to how a computer's circuits operate with electrical impulses. However, the brain is massively parallel and plastic, whereas computers are serial and fixed architecture."

Example 2 – A second dialogue with an LLM

Example n.2 shows a decent (if somewhat superficial) analogy and a commonly used one. Does the LLM truly understand the brain or computers at a deep level? No, it is relying on known analogies from its training data. In 2023, there was debate over whether such analogical reasoning by LLMs is genuinely robust or merely superficial. Yuan et al. (2023) introduced the SCAR benchmark to evaluate structural (relational) analogies. They found that LLMs (including GPT-4) often struggle with tasks that require deeper analogical insight. For example, an LLM might know the common (and incorrect) analogy "atom is like a solar system" with electrons as planets, etc.. But, if given a new analogy puzzle outside its training, it may rely on shallow pattern matching, often failing to maintain the correct structural alignment. This indicates that while LLMs can generate analogies, they may lack the systematicity required for genuine analogical reasoning. This implies that the explanations LLMs produce may also lack systematic rigour. They sound convincing because they imitate familiar explanatory patterns, but they may omit subtle conditions or caveats that a rigorous human reasoner would include. In Example n. 2, the LLM did mention a disanalogy (parallel versus serial), which is positive. But that is also something it has seen stated; it does not derive it anew.

At this point, one might ask: is the resemblance between LLM outputs and human explanations merely an illusion for the observer, or does it indicate that LLMs implicitly perform some form of inference? Some cognitive scientists argue that LLMs have learned representations that encode aspects of the world sufficiently to perform implicit multi-step inferences—such as multi-hop question answering (questions requiring the combination of two facts)—even if not done through formal logic. Larger models handle multi-hop questions more effectively than smaller ones, suggesting emergent capabilities beyond one-step pattern matching. This presents a continuum: LLMs are not merely "dumb" parrots; they possess generalisation abilities that enable them to recombine known pieces in novel ways. To humans, this may seem a rudimentary form of reasoning, but it is better characterised as statistical inference, which is not guaranteed to be correct but often yields sensible conclusions. Only metaphorically could one say that LLMs use a form of abductive heuristics. They have seen many problems and their solutions, so when faced with a new problem, they appear to reach a solution that would best explain the problem if it were an instance of something they "know" (Imran et al., 2025). Sometimes this heuristic hits the mark, sometimes it does not.

**6. Objections**

Our argument may face several objections. We discuss five here, which we believe are more significant, to clarify the scope and limitations of our claims.

*Objection 1* (based on Mirzadeh et al. 2025, Shojaee et al. 2025, Zhao et al. 2025). LLMs don't really perform inference, so comparing them to abduction/IBE is misguided.

*Reply 1*. According to this view, any appearance of reasoning in LLMs is an illusion for the user, and invoking philosophical concepts like abduction risks anthropomorphising the model (Floridi & Nobre 2024). Indeed, scholars warn against "unreflectingly apply[ing] to AI systems the same intuitions that we deploy in our dealings with each other" (Shanahan 2022). An LLM does not possess beliefs, intentions, understanding, propositional attitudes, or mental states, so can we really say it "infers" anything? We are not claiming that LLMs hold literal beliefs or follow Peirce's method of hypothesis internally. Instead, we argue that the output structure of LLMs often resembles that of an abductive reasoning process, which is frequently reinforced by interface design, and this resemblance is not random but systematic, resulting from training on human explanations. When an LLM generates a plausible hypothesis for data, it is reasonable to draw an analogy to abduction, as long as we acknowledge it as an analogy. The stochastic generation process can be viewed as exploring the space of possible continuations, guided by learned probabilities. In that sense, it performs an inference: choosing the next word that best continues the text. This "inference" is purely statistical, but because human-like reasoning is embedded in the statistical patterns, the outcome can be mapped onto reasoning.

*Objection 2* (based on Bubeck et al. 2023, OpenAI, 2023, Webb et al. 2023, Lewis & Mitchell 2025, Li et al. 2025). If LLMs are just stochastic parrots, why do they sometimes outperform humans on reasoning tasks? For example, GPT-4 has shown high accuracy on specific professional exams and logical puzzles. Webb et al. (2023) even reported that GPT-3 and GPT-4 matched or surpassed human performance on some abstract analogy problems. Does this contradict the idea that LLMs lack real reasoning?

*Reply 2.* No, when an LLM surpasses humans on a task, it could be because it has encountered many examples during training and has effectively learned patterns that humans might find unintuitive. For instance, the excellent performance of GPT-4 on the Raven's Progressive Matrices (a visual-analogy test) in the Webb et al. study may be due to textual descriptions of such problems or to the underlying patterns present in its training data (or to fine-tuning to succeed on them). It is also possible that, due to their extensive training, LLMs develop a kind of ensemble effect: they combine various approaches found in their data, which can make them unexpectedly robust on some tasks (similar to how tearing a piece of paper is easy, but the thicker the stack, the more difficult it becomes). Nonetheless, this does not amount to understanding; it is more like a student who has seen many example solutions and can pattern-match to solve a new problem in the same format. But if the format is slightly altered, for example, a puzzle with a twist unlike any training example, humans can adapt while the LLM may fail. For example, a model that solves arithmetic word problems phrased in one way but struggles if phrased differently, indicating it did not comprehend or understand—or whatever term one prefers to mean "get"—the underlying arithmetic reasoning, but rather the template of the question. Furthermore, when carefully evaluated, LLMs still make reasoning errors that humans with basic training usually avoid. For example, they can be misled by logical fallacies or produce inconsistent results. One study (Payandeh et al. 2024) tested GPT-3.5 and GPT-4 on various logical fallacies and found that the models often accepted fallacious reasoning unless explicitly prompted to critique it. While humans are also vulnerable to fallacies, they can be trained to recognise them. This indicates that LLMs lack a metacognitive check on the consistency of their reasoning; they can generate both a statement and its negation in different contexts without realising that they conflict. In summary, sporadic outperformance is not proof of genuine reasoning ability; it is often a sign of overfitting to common patterns.

*Objection 3* (based on Zhou et al. 2023, Yamin et al. 2024, Wu et al. 2025). Abductive reasoning in humans involves common sense and causality; LLMs have none, so how can their outputs truly resemble abduction beyond superficial word patterns?

*Reply 3.* A solid foundation of causality and physical understanding indeed supports human reasoning. We see smoke and infer the presence of fire because we know that fire typically causes smoke. Large Language Models (LLMs) see the word "smoke" and often output "fire" because, in their training data, these words frequently co-occur as cause and effect. The difference is subtle: humans infer real fire in the world, while LLMs predict "fire" within sentences. However, if asked, "There is smoke. What is a possible cause?", it will answer "Fire" in a causal sense, not just to complete a sentence, because it has learned that causal relation as a linguistic association. Essentially, the LLM's extensive training on language has endowed it with a vast repository of commonsense causal knowledge, although not explicitly structured. It "knows" that slippery floors cause falls, that not eating causes hunger, that polls predict elections, and so on—because it has processed countless expressions of these relations. This enables it to perform some degree of commonsense reasoning. What it lacks, however, is an experiential or embodied grounding of that knowledge. It does not have sensorimotor verification, such as pushing a cup off a table and causing it to fall. But in language, it has likely encountered "the cup fell off the table after being pushed," which associates "push" with "fall." One may argue that this purely text-based knowledge is fragile and incomplete. That's a fair point: linguistic associations alone may miss real-world nuances. For example, the model might not understand gravity as a universal law, but only through specific anecdotes. Nevertheless, the success of LLMs in answering many causal questions indicates that the statistical abstraction of cause-and-effect in the training data is often sufficient to mimic human causal reasoning.

*Objection 4* (based on Lauriola et al. 2025, Zheng et al. 2024, Cao et al. 2024). Your analysis is too generous; aren't LLM outputs often incoherent or irrelevant, nothing like good explanations?

*Reply 4*. LLM outputs can indeed decline in quality, especially when trained on low-quality data, with smaller models, or when given poor prompts. They may diverge from the main topic, miss the intent of a question, or generate generic responses. Not every output resembles a precise IBE; sometimes, it feels like nonsense. Often, it mirrors standard platitudes. Our analysis has focused on situations where LLMs succeed in delivering explanation-like answers. However, it is essential to remember that this requires a sufficiently capable model and often requires careful prompting. An LLM might produce a mediocre response in a zero-shot, unprompted setting. For instance, ask a straightforward riddle, and a smaller model might stumble or give a nonsensical answer, while a human would reason it out. These failures remind us that stochastic pattern matching does not ensure coherence: it can latch onto incorrect patterns. That said, top-tier models guided by instructions have significantly reduced incoherence, to the extent that many answers seem thoughtfully composed. The fact that coherence varies with model quality highlights our core premise: nothing extraordinary has been added to each new version in the GPT series, apart from scale and the breadth of training data, which enhance the statistical approximation of language. With ample data and parameters, the model captures more of the coherence present in human discourse. While an earlier GPT might have provided an explanation that was somewhat relevant but partly mistaken, GPT-5's explanation is likely to verify more points. This progress suggests that the "abductive appearance" is more than mere coincidence. As the model better captures human language regularities, its explanations become increasingly indistinguishable from those crafted by humans. Nevertheless, limitations remain, even for the best models, such as difficulty with specialised or complex cases that humans can handle with insight. We are not claiming that LLMs match human reasoning abilities. Instead, they project an imitation of a large subset of it, and this projection becomes more convincing as models improve.

*Objection 5*. There remains one more objection that seems implicit in the literature. Your analysis is limited to LLMs insofar as they are based on the token-completion paradigm. What about other kinds of LLM?

*Reply 5*. This objection can be answered in three steps. First, we anticipated that we were addressing current, mainstream LLMs based on the token-completion paradigm, such as the GPT series. This may seem an admission of "temporality": our analysis may be made outdated by new approaches. Of course, we cannot rule this out. We are addressing current technology. However, the second step is to note that, at the time of writing, the most successful model in the GPT series, GPT 5.1, remains a token completion model. Like all models in the GPT series, it predicts and generates the next sequence of tokens based on the input prompt and the ongoing conversation context. Importantly, its "reasoning" capability does not fundamentally distinguish it from a token completion model; rather, it is an advanced feature implemented using the token completion mechanism itself. The model generates internal, hidden tokens that function as a scratchpad before producing the final user-facing output, essentially prompting itself to "think step by step" or "outline a plan" internally. The tokens generated during this internal process are not shown to the user immediately but help the model follow complex instructions, perform multi-step logic, and reduce errors (hallucinations). Third, if future token-completion models incorporate some form of "abduction engine," it would only reinforce our point. Current LLMs are stochastic engines rather than abductive ones, to the point that genuine abduction requires augmenting them. By analogy, consider mathematical computation: LLMs are not designed to serve as advanced calculators. Precisely because they are not reliably accurate at complex math (owing to fundamental differences between how they process language and how computers execute mathematical operations), modern systems often integrate actual calculator tools or code interpreters to achieve highly accurate results.

## 7. Limitations

Having addressed the previous objections, we must recognise some limitations in our own analysis. We have treated "LLMs" somewhat generally, focusing mainly on the latest large models as of 2025, with the GPT series as a reference point. Smaller or less trained models might not exhibit the abductive illusion as strongly; their outputs can be clearly incorrect. We remain agnostic about future possible systems. Furthermore, our epistemological level of abstraction (stochastic versus abductive) may not capture all nuances. There are other reasoning forms and aspects of LLMs, such as memory and attention mechanisms, that we have not explored here. Someone could argue that we have not rigorously defined "explanation" either, as we have used it in a commonsense way. In philosophy of science, explanation is theorised in different ways, which we have not applied here.[12] Doing so could be enlightening: for example, does an LLM's explanation meet any formal criteria of explanation, like unification or causation? Likely not explicitly, but perhaps implicitly, it often aligns with a causal model because language encodes causal information. These are areas for further exploration. Additionally, we have not truly engaged with ethical or epistemological implications, such as trust or the value of explanations from a model when it does not "know" whether they are correct. These issues are important: if a model provides a convincing but incorrect explanation, users may trust it unnecessarily, thereby spreading misinformation. That is arguably an epistemic harm of the abductive illusion: it can mislead us into treating informed guesses as if they were knowledge. Critics (Lenat & Marcus, 2023) have highlighted this point, noting that without proper understanding or grounding, LLM explanations are unreliable and potentially dangerous, if used unchecked in domains like medicine or law. Our analysis agrees and provides the conceptual foundation: the LLM is not truly reasoning towards the best explanation; it is reproducing a probable explanation. Therefore, one should regard its output more as the opinion of an anonymous forum poster—possibly correct, possibly incorrect—rather than an expert.

## 8. Conclusion: Stochastics at the Core, Abduction on the Surface

Abductive reasoning in AI is an active area of research (Yang et al. 2023, Abdaljalil et al., 2025), but what is the relationship between the stochastic nature of LLMs and their tendency to produce seemingly abductive, explanatory outputs? We have argued that the core of LLM behaviour lies in stochastic pattern learning, yet their outputs resemble abductive reasoning. Fundamentally, an LLM is a number-crunching system that uses vast statistics of language usage (token distributions, co-occurrences, sequence likelihoods) to generate text. It knows nothing in the ordinary sense of "knowledge" (Reichenbach, 1938; Searle, 1980). It proves nothing; it does not follow the rules of inference or logic. In Peirce's terms, it performs no logical energy; it is entirely a pattern "habit." Nevertheless, when we interact with an LLM, it feels like engaging with a reasoning agent. It offers explanations, analogies, and even "judgments" about what is likely or essential. This appearance arises from the model being fed the products of human thought and, of course, the clever, if tricky, design of relevant interfaces. A necessary clarification is that understanding LLMs involves distinguishing their internal processes from their outputs. Internally, it involves random sampling guided by probabilities; externally, it can produce answers that align with human reasoning norms. Therefore, we can briefly describe LLMs as fundamentally stochastic, with surface-level abductive appearances. Recognising this duality helps clarify some debates: we can agree with sceptics that no human-like understanding occurs internally, while also explaining why these models are so successful and attractive: they leverage the informational richness of human language and thus effectively stand on the shoulders of our collective knowledge and reasoning.

---

[12] For a good overview, see (Woodward, 2003).

This insight holds both promise and danger. On the one hand, it places LLMs within the long history of logic and statistics: we see them not as new forms of intelligence or alien minds, but as continuations of a trajectory where formal methods are used to model aspects of human thought, implemented through computational systems capable of acting as agents. They represent the novelty of "engines of generative plausibility": never before have we had systems capable of producing human-like, plausible text at scale. This opens up opportunities: these engines can draft explanations, brainstorm hypotheses, translate complex information into simpler language, and more. They could serve as educational tools—explaining concepts on demand and at appropriate levels of complexity and education—or as tools to enhance creativity. In fields such as medicine or law, they could quickly suggest possible explanations or solutions for a human expert to review. In science, they might scour literature and propose theoretical connections. They could help us challenge preconceptions and implicit orthodoxies. They are the best interfaces we have today for accessing, querying, and managing the immense accumulation of human content. All this relies on using surface-level abductive cues wisely, while compensating for the stochastic core's unreliability and lack of understanding.

On the other hand, the dangers are clear: if one conflates the surface with the core—if one assumes the LLMs genuinely "know what they are talking about"—one can be misled. We have already seen instances of chatbots persuading users of false or harmful ideas by sounding authoritative. The veneer of explanation can be a trap. Philosophically, this raises questions about the difference between explanation and truth in LLMs' outputs. An explanation can be coherent and convincing (even optimal by IBE criteria) and yet still false (Pettigrew 2022). LLMs lack an epistemic compass to navigate that distinction. As users or deployers of LLMs, we must provide that compass externally (He et al., 2024), for example, through human oversight or new system architectures, such as integrating LLMs with knowledge graphs, verification modules, search engines, and RAG, or by setting the right "temperature" and an ability to say "I do not know". These powerful tools have increased our epistemic responsibilities.

Zheng, C., Zhou, H., Meng, F., Zhou, J., & Huang, M. 2024. Large language models are not robust multiple choice selectors. In Proceedings of the 12th International Conference on Learning Representations (ICLR 2024).

Zhou, Y., Wu, X., Huang, B., Wu, J., Feng, L., & Tan, K. C. 2023. Large language models may talk causality but are not causal. Transactions on Machine Learning Research.26